\RequirePackage{fix-cm}
\documentclass{svjour3}                     
\smartqed  
\usepackage{graphicx}
\usepackage{url} 
\usepackage{xcolor}
\begin{document}
\title{Point detection through multi-instance deep heatmap regression for sutures in endoscopy
\thanks{The research was supported by the German Research Foundation DFG Project 398787259, DE 2131/2-1 and EN 1197/2-1 and by Informatics for Life funded by the Klaus Tschira Foundation}
}

\titlerunning{Multi-instance Point Detection}        

\author{Lalith~Sharan         \and
        Gabriele~Romano     \and
        Julian~Brand   \and
        Halvar~Kelm \and
        Matthias~Karck        \and
        Raffaele~De~Simone   \and
        Sandy~Engelhardt 
}



\institute{L. Sharan \and J. Brand \and H. Kelm \and S. Engelhardt\at
              Department of Internal Medicine III, Group Artificial Intelligence in Cardiovascular Medicine, Heidelberg University Hospital, D-69120 Heidelberg, Germany \\
              \email{lalithnag.sharangururaj@med.uni-heidelberg.de,}          
              \email{sandy.engelhardt@med.uni-heidelberg.de}%
\and
 G. Romano \and M. Karck \and R. De Simone \at
              Department of Cardiac Surgery, Heidelberg University Hospital, D-69120 Heidelberg, Germany }

\date{Received: date / Accepted: date}

\maketitle

\begin{abstract}
\textit{Purpose:} Mitral valve repair is a complex minimally invasive surgery of the heart valve. In this context, suture detection from endoscopic images is a highly relevant task that provides quantitative information to analyse suturing patterns, assess prosthetic configurations and produce augmented reality visualisations. Facial or anatomical landmark detection tasks typically contain a fixed number of landmarks, and use regression or fixed heatmap-based approaches to localize the landmarks. However in endoscopy, there are a varying number of sutures in every image, and the sutures may occur at any location in the annulus, as they are not semantically unique. 
\textit{Method:} In this work, we formulate the suture detection task as a multi-instance deep heatmap regression problem, to identify entry and exit points of sutures. We extend our previous work, and introduce the novel use of a $2D$ \textit{Gaussian} layer followed by a differentiable $2D$ spatial \textit{Soft-Argmax} layer to function as a local non-maximum suppression. 
\textit{Results:} We present extensive experiments with multiple heatmap distribution functions and two variants of the proposed model. In the intra-operative domain, Variant $1$ showed a mean $F_1$ of $+0.0422$ over the baseline. Similarly, in the simulator domain, Variant $1$ showed a mean $F_1$ of $+0.0865$ over the baseline.
\textit{Conclusion:} The proposed model shows an improvement over the baseline in the intra-operative and the simulator domains. The data is made publicly available within the scope of the MICCAI AdaptOR2021 Challenge \url{https://adaptor2021.github.io/}, and the code at \url{https://github.com/Cardio-AI/suture-detection-pytorch/}.

\keywords{Point Detection \and Mitral Valve Repair \and Endoscopy}
\end{abstract}

\section{Introduction}
\label{introduction}

Mitral valve repair is a surgery of the mitral valve of the heart, that seeks to restore its function by reconstructing the valvular tissue. In this surgery, a prosthetic ring is affixed to the mitral valve, by first suturing around the annulus of the valve and then implanting the ring of a chosen size, through the sutures onto the annulus \cite{carpentier_new_1971}. Mitral valve repair is increasingly performed in a minimally invasive manner \cite{casselman_filip_p._mitral_2003}, with a reliance on image guidance, in particular endoscopic video for the reconstruction process. 

Besides, the use of surgical simulators are becoming more popular in training and familiarizing the surgeons with the demanding surgical techniques. In our previous work, we showed how to simulate endoscopic surgeries on a patient-individual basis with the help of flexible $3D$-printed mitral valve replica \cite{EngelhardtIJCARS2019,EngelhardtICVTS}. 
The endoscopic data stream obtained during surgery or such simulations can be analysed in real-time or retrospectively to extract quantitative information with regards to patient-valve geometry \cite{sharan_domain_2020} or context-aware visualisations.

In particular, suture detection is one such task that can provide quantitative information. This information can then be used to analyse the suturing patterns, examine the correlation with different levels of expertise, understand the optimal suture configuration in the context of ring implantation, and to use the suture locations to create augmented reality visualisations \cite{EngelhardtMiccai14}. The task of suture detection entails detecting the entry and exit point of the sutures around the annulus. More precisely, given an image and the corresponding suture locations for this image, the task is to predict the suture locations for unseen endoscopic images. There have been multiple approaches from the literature in the field of landmark detection, more commonly in facial landmark detection \cite{yan_survey_2018}, pose estimation \cite{bulat_human_2016,iqbal_hand_2018} and medical landmark detection \cite{zhang_detecting_2017,sofka_fully_2017} to tackle this problem. However, there are two important distinctions due to which these approaches are not directly applicable to our task. Firstly, for any given image there exists a variable number of sutures, unlike a fixed number of facial or anatomical landmarks. Secondly, the points have a semantic meaning and are of single-instance. This renders fixed regression-based approaches ill-suited to our task. Additionally, commonly used patch-based refinement of anatomical landmarks are inapt in this scenario. 

In this approach, we seek to solve a multi-instance detection problem for 2D points. The approach is based on a heatmap which typically models the distribution of likelihood around the point as a Gaussian. 
In this paper, we extend on our previous conference work \cite{DBLP:conf/bildmed/SternSRKKSWE21}, where we proposed a simple U-net with a single channel output. The output map was thresholded and the center of mass was calculated for each region to determine the final position of points. 

In the work at hand, we introduce the usage of a differentiable \textit{Gaussian} filter and a \textit{Soft-Argmax} layer to enforce both, learning of the heatmap and further extracting the points from the heatmap, in a differentiable manner. We demonstrate an improvement compared to our previous baseline, and additionally perform experiments comparing various final layer configurations and loss functions. The approach is evaluated on two different domains, i.e. video data from simulated surgery and from real procedures are used. The data is made publicly available within the scope of the MICCAI AdaptOR2021 Challenge \url{https://adaptor2021.github.io/}. To be consistent with the challenge, we used the same data set split like in the challenge, which is slightly different from \cite{DBLP:conf/bildmed/SternSRKKSWE21}.

\section{Related Work}
\label{related_work}

\subsection{Regression-based approaches}
\label{regression}
Discriminative approaches to landmark detection comprise of regression or heatmap based methods. Regression based methods directly estimate the landmark coordinates from the image. Duffner and Garcia \cite{duffner_connexionist_2005} is one of the early works using neuronal layers to estimate facial feature positions. Subsequently, due to the exponential growth of deep learning tools and techniques, there have been a number of works that estimate this mapping with a neural network \cite{yan_survey_2018}. 
The regression based methods model a mapping from the image space to the coordinate space, which is highly non-linear and therefore difficult to learn. The approaches from the literature tackle this problem by using a \textit{CNN} cascade or progressive refinement of the landmarks \cite{fan_approaching_2016,yang_stacked_2017}. 

In the field of medical imaging, landmark localization is a relevant step for tasks such as registration and augmented reality visualisations. Cascaded and stage-wise models \cite{DBLP:journals/corr/SunMX15,zhang_detecting_2017} are typically used for anatomical landmark regression.
Sofka \textit{et al.} \cite{sofka_fully_2017} presented a landmark regressor for ultrasound image sequences, that additionally imposes a temporal constraint with \textit{LSTM} cells along with a center-of-mass layer to extract landmark locations. 
However, all of the aforementioned methods predict a pre-defined number of landmarks, which allows to pre-determine the shape of the tensor to be regressed in the output layer of a network. Additionally, learning a transformation from the image space to the coordinate space is a highly non-linear mapping which is further complicated by the variations in camera view, pose, illumination and scene composition. 

\subsection{Heatmap-based approaches}
\label{heatmap}
Unlike regression based approaches that directly regress on the coordinates, heatmaps model a distribution of likelihood around the points of interest. In the recent years, the field of landmark detection is moving towards the use of heatmap based approaches \cite{yan_survey_2018}, as they model a mapping from image-to-image space unlike regression models. A \textit{Gaussian} distribution is commonly used to model the likelihood of the landmark locations. Typically, the heatmap approaches represent a single landmark in one channel, making it easy to perform differentiable operations or post-processing \cite{Chandran2020AttentionDrivenCF}. Bulat and Tzimiropoulos \cite{bulat_human_2016} proposed a two-stage network to regress on heatmaps and further finetune the landmarks in subsequent stages. The \textit{Deep Alignment Network (DAN)} \cite{kowalski_deep_2017} processes the whole image in contrast to patches and finetunes the landmark estimates using heatmaps. In the medical domain, Zhang \textit{et al.} \cite{DBLP:conf/miccai/ZhangLWCYLSTCXS17} used a multi-task network to learn displacement maps using heatmaps. Payer \textit{et al.} \cite{payer_integrating_2019} proposed a two stage heatmap based network. 
All in all, similar to the regression approaches, the heatmap based regression networks, model one heatmap per channel with a pre-defined set of landmarks. 

In earlier works on a small intra-operative dataset \cite{EngelhardtMiccai14,3093-04}, we have used random forests and tailored post-processing for point detection and optical flow for point tracking. Our previous work on the same data base  \cite{DBLP:conf/bildmed/SternSRKKSWE21} formulates the landmark detection task as a deep learning based approach, and demonstrates first results on intra-operative and surgical simulator datasets for heart surgeries. Hervella \textit{et al.} \cite{DBLP:journals/cmpb/HervellaRNPO20} demonstrated a similar method for the case of retinal fundus images. However, the model has to learn from a heavily unbalanced dataset due to the nature of point landmarks in the context of a segmentation task. Brosch \textit{et al.} \cite{brosch_deep_2015} tackles this problem by using a novel objective function. In this paper, we extend our previous work \cite{DBLP:conf/bildmed/SternSRKKSWE21} and tackle the unbalanced multi-instance sparse-segmentation task through the use of a differentiable convolutional \textit{Soft-Argmax} layer combined with a balanced loss function. Iqbal \textit{et al.} \cite{iqbal_hand_2018} used a differentiable \textit{Soft-Argmax} layer to extract the landmark locations from the heatmap, but the problem formulation contained a single heatmap per channel for a pre-defined number of heatmaps. Chandran \textit{et al.} \cite{Chandran2020AttentionDrivenCF} used a heatmap combined with the differentiable \textit{Soft-Argmax} layer to extract regions of interest to provide a global context to landmark localization. In contrast to Iqbal \textit{et al.} \cite{iqbal_hand_2018} and Chandran \textit{et al.} \cite{Chandran2020AttentionDrivenCF}, our approach uses a convolutional \textit{Soft-Argmax} layer that is convolved spatially with the feature map, in order to impose stability in modeling a distribution and extracting multiple instances of landmarks from this distribution. Additionally, an $F_\beta$ loss is used to take into account the precision and recall for optimisation. Results compared to our previous baseline \cite{DBLP:conf/bildmed/SternSRKKSWE21}, along with ablations with various loss functions and output layer configurations, are presented.

\section{Methods}
\label{methods}

\subsection{Task Formulation}
\label{task_formulation}
\begin{figure*}
  \includegraphics[width=1\textwidth]{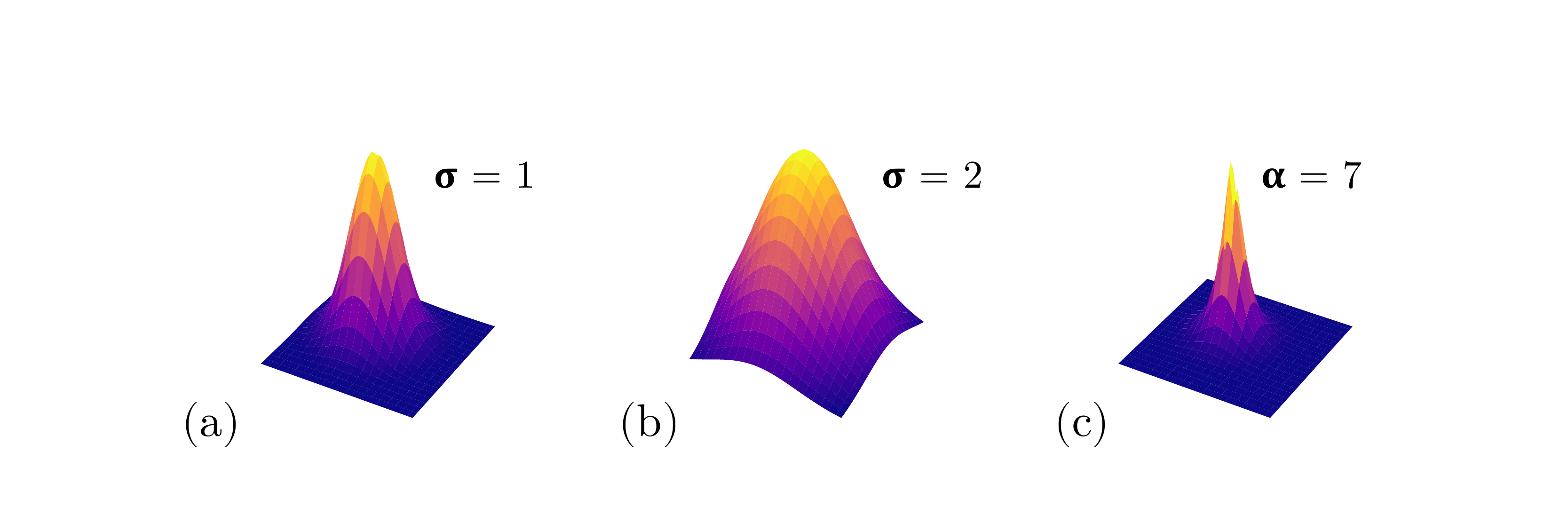}
\caption{Distribution functions. (a) \textit{Gaussian}, $\sigma=1$ (b) \textit{Gaussian}, $\sigma=2$ (c) \textit{Tanh}, $\alpha=7$ }
\label{fig:distributions}
\end{figure*}

The labels $s_i \in S, i \in 1, 2....N$ can be equivalently represented as a binary mask, where $p_{(x_i, y_i)} \in \{0, 1\}, x_i \in 1, 2....,H, y_i \in 1, 2....,W$, denotes the pixel value at image location $(x_i, y_i)$, with a value of $1$ in the suture locations and $0$ otherwise. Alternatively, the position of each suture instance can be modelled by a distribution centered around the location that represents the likelihood of the pixel being a landmark location. In this case, $p_{(x_i, y_i)}$ takes values in $[0, 1]$. In this work, we present and compare two different distribution functions to model the heatmap, namely the \textit{Gaussian} and the \textit{Tanh} distribution. 
In the case of the \textit{Gaussian} distribution, the spread is controlled by the variable $\sigma_1$, which we set to $\sigma_1=1$, $2$, and $3$. The \textit{Tanh} distribution is a sharper distribution, where we set the variable $\alpha=3.5\times\sigma$ that controls the spread of the distribution. We experiment with the values of $\alpha=7$ and $\alpha=10.5$. An illustration comparing the \textit{Gaussian} and \textit{Tanh} distribution is provided in Fig. \ref{fig:distributions}.

\subsection{Network Architecture}
\label{architecture}
The labels $\hat{y}_{j}$ for unseen endoscopic images can be estimated by a neural network $\phi(x_j, y_j, \theta)$ with parameters $\theta$. A U-Net \cite{ronneberger_u-net_2015} based architecture is used, similar to the one described in our previous work \cite{DBLP:conf/bildmed/SternSRKKSWE21}, but using \textit{ReLU} activations for the convolutional layers. 
\textit{RGB} $3$-channel images of size $288 \times 512$ are provided to the model as input. A mask of the same size is created from the labelled suture locations. A distribution function centered around each suture point is applied to the binary mask as described in Section \ref{task_formulation}. 

\begin{figure*}
  \includegraphics[width=1\textwidth]{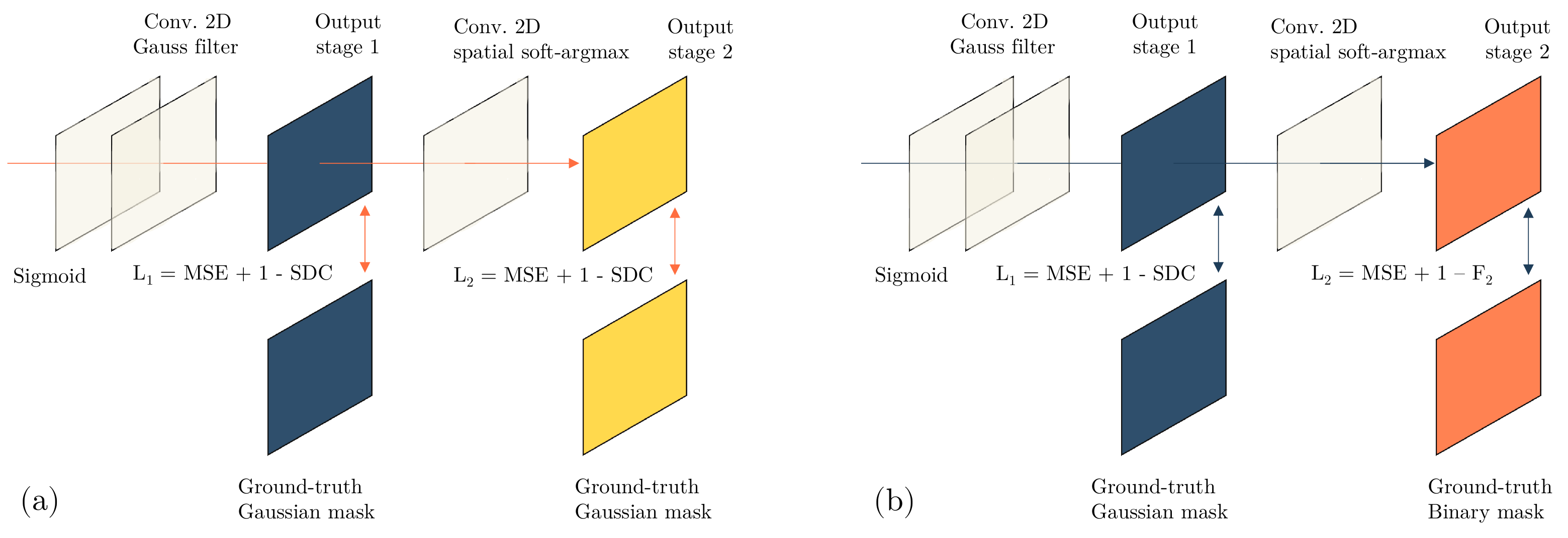}
\caption{Overview of the variants for the proposed suture detection network. (a) Variant $1$: A \textit{Gaussian} mask is used at both output stages. (b) Variant $2$: A binary mask is used at output Stage $2$.}
\label{fig:architecture}     
\end{figure*}

Furthermore, a differentiable \textit{Gaussian} filter with spread $\sigma_2$ is applied to the output of the \textit{Sigmoid} layer. Different values of $\sigma_1$ and $\sigma_2$ are applied and a comparison is presented in Section $5.1$. A similarity loss $\mathcal{L}_{1}$ between the filtered output (\textit{Output Stage} $1$ in Fig. \ref{fig:architecture} (a) and (b)) and the ground-truth heatmap is applied, that enforces the model to learn the likelihood distribution of the suture locations. The \textit{Gaussian} filter encourages the model to learn a smooth distribution around the predicted locations. Additionally, the filtered output is fed through a differentiable convolutional $2D$ spatial \textit{Soft-Argmax} layer to produce the final output of the model (\textit{Output Stage} $2$ in Fig. \ref{fig:architecture} (a) and (b)). In this layer, a \textit{Soft-Argmax} kernel of size $(3 \times 3)$, with a stride of $1$ and a padding of $1$ is convolved with the output from the previous layer. The layer is implemented using the \textit{Kornia} \cite{eriba2019kornia} library. An additional similarity loss $\mathcal{L}_{2}$ is applied at this stage, as illustrated in Fig. \ref{fig:architecture} (a) and (b). Other works  \cite{iqbal_hand_2018,Chandran2020AttentionDrivenCF} demonstrated the use of a differentiable \textit{Soft-Argmax} layer in extracting the landmarks from the heatmaps, where a single landmark is used per channel of the heatmap and as a result, the \textit{Soft-Argmax} layer yields the landmark locations. In contrast, we represent all  suture locations in a single channel. Therefore, the convolutional \textit{Soft-Argmax} layer functions as a local non-maximum suppression for the points with low likelihood of being a suture point. 

At the output of stage $1$, a loss function of the form $\mathcal{L}_{1} = MSE + 1 - SDC$ is applied between the predicted suture points and the ground-truth heatmap, where $MSE$ is the \textit{Mean Squared Error} and SDC is the \textit{Sørensen Dice Coefficient}. For the model variant $1$ at the Output Stage $2$ as shown in Fig. \ref{fig:architecture} (a), the same loss $\mathcal{L}_{2} = MSE + 1 - SDC$ is applied for a heatmap ground-truth. For the model variant $2$ at the Output Stage $2$ as shown in Fig. \ref{fig:architecture} (b), the output is optimised with a binary ground-truth mask. In this case, the true and false pixel classes are even more unbalanced. An asymmetric similarity loss function that can weight the precision and recall, is previously shown to perform better for unbalanced classes \cite{hashemi_asymmetric_2019}, in comparison to the dice coefficient. We therefore apply a balanced $F_\beta$ loss function for the predicted and ground-truth binary masks

\begin{equation}
F_\beta = \frac{{(1+\beta^2)}{\sum_{i=1}^{N} {p_i}{g_i}}} {{{(1+\beta^2)}{\sum_{i=1}^{N} {p_i}{g_i}}}
+{{\beta^2}{\sum_{i=1}^{N} {(1-p_i)}{g_i}}} + {{\sum_{i=1}^{N} {p_i}{(1-g_i)}}}} 
\label{eq:f_beta_loss}
\end{equation}

where $p_i$ is the likelihood of a pixel being a suture, and the binary label $g_i \in \{0,1\}$ denotes the presence of a suture point. 
A value of $\beta=2$ is used, to penalise the number of false negatives.
The final suture detection model, jointly optimises the loss functions $\mathcal{L}_{1}$ and $\mathcal{L}_{2}$, given by

\begin{equation}
\mathcal{L}_{\phi(x_j, y_j, \theta)} = \min_\theta(\mathcal{L}_{1} + \mathcal{L}_{2}). 
\label{eq:f_model_equation}
\end{equation}

The predicted heatmaps that are obtained after evaluation are thresholded with $t=0.5$, and the center-of-mass of the local clusters are computed, to extract the predicted suture coordinates, that are then evaluated with the labeled suture coordinates.

\subsection{Evaluation}
\label{evaluation}
\begin{figure*}
  \includegraphics[width=1\textwidth]{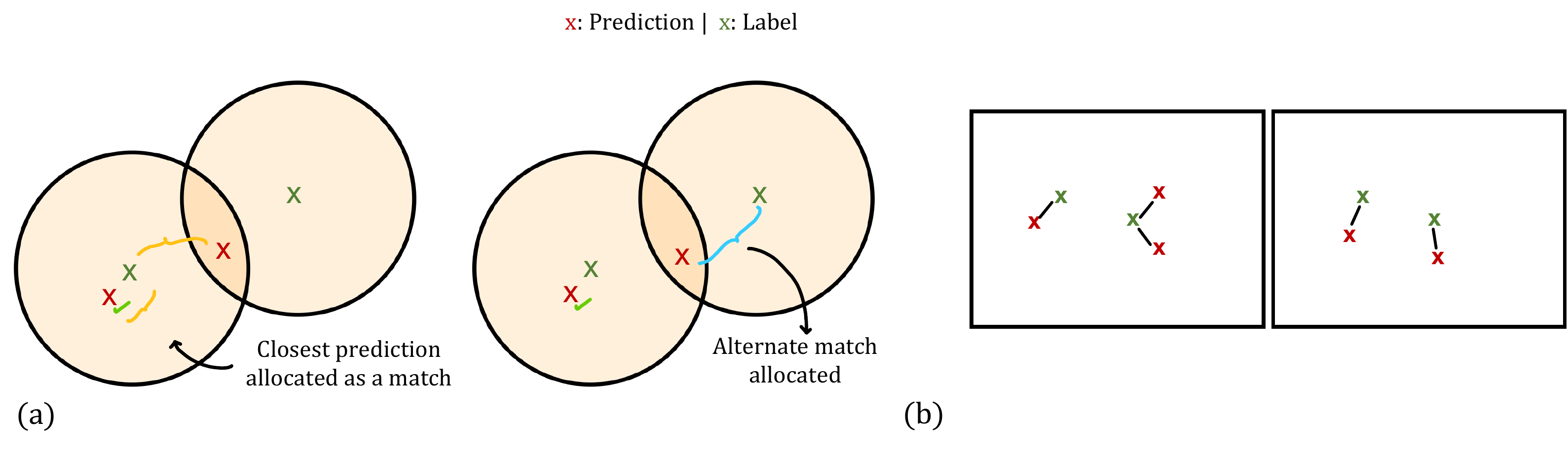}
\caption{(a) An illustration of allocating matches between the predicted and ground-truth suture labels. (b) A comparison of cases that yield a similar $RMSE$ metric.}
\label{fig:evaluation}     
\end{figure*}

 A suture point is considered to be successfully predicted, if the distance between the predicted and ground-truth point is less than $6$ pixels, as proposed in \cite{DBLP:conf/bildmed/SternSRKKSWE21}. If multiple points are matched with the ground-truth, then the point closest to the ground-truth is chosen as the predicted point. Once the closest match is allocated the second-best match is used to match other labels in the image. An illustration of this is shown in Figure \ref{fig:evaluation} (a). From these predicted co-ordinates, the number of \textit{True Positives (TP)}, \textit{False Positives (FP)}, and \textit{False Negatives (FN)} are determined. Furthermore, the \textit{Positive Predicted Value (PPV)} is computed as $PPV = TP/(TP + FP)$ and the \textit{True Positive Rate (TPR)} is determined as $TPR = TP/(TP + FN)$. In order to compare the performance of two different models, the $F_1$ score is computed by taking the harmonic mean of $PPV$ and $TPR$ as $F_1 = (PPV \times TPR) / (PPV + TPR)$. Additionally in this work, we compute the root mean square error of the Euclidean distance as a localisation metric. For each predicted point in an image, the Euclidean distances to all ground-truth points are computed and the least distance is chosen. This is then averaged across all predicted points in an image, for the images that have predictions. For images without predicted points, the metric cannot be computed, and we therefore additionally report the number of images where this occurs in Table \ref{tab:metric_distance}. For $RMSE$ computation, we consider each predicted point in the calculation, i.e. each point has a match. Points which are further away without a match would otherwise not be penalized in the metric.
 
\section{Data and Experiments}
\label{data_experiments}

\subsection{Datasets}
\label{datasets}
In this work, datasets from two domains are used for the experiments, namely: the intra-operative, and the surgical simulator domain. The data in the intra-operative domain comes from endoscopic frames captured during mitral valve repair surgery. The intra-operative data forms a heterogeneous dataset comprising of frames with widely varying camera view, scale, lighting sources, and white balance. Additionally, the dataset also contains the presence of endoscopic artefacts caused due to occlusion and specular reflections. The intra-operative datasets are split on a surgery level, like in the mentioned Challenge. Firstly, a dataset for cross-validation ($A.1$) is created, comprising of $4$ surgeries, with a total of $2,376$ images. It is important to note, that the validation set is not used for fine-tuning the model performance, for example, using early stopping. Therefore, the validation set functions as an unseen dataset for the model. Additionally, a second independent intra-operative test set is created ($A.2$). The data in the surgical simulator domain comes from endoscopic capture of the surgical training and planning sessions on the mitral valve silicone replica models \cite{EngelhardtICVTS}. A simulator dataset ($B.1$) is created from $10$ such simulator sessions, with a total of $2,708$ images. 

The endoscopic videos are captured in a top-down stereo format, after which relevant frames are extracted. The left and right images of the stereo pair are treated independently. The suture points identified in these frames are then manually labeled using the annotation tool \textit{labelme}. For further details about the endoscopic capture of the data, the reader is referred to our previous work \cite{DBLP:conf/bildmed/SternSRKKSWE21,9496194}. For this work, the previously annotated suture point labels are revisited and quality-checked. 
After correction, the intra-operative cross-validation dataset contains a total of $23,938$ sutures, and the simulator dataset a total of $33,872$ sutures.The data is released within the scope of the AdaptOR2021 MICCAI challenge at \url{https://adaptor2021.github.io/} 

The endoscopic frames are resized to a resolution of $288 \times 512$, and image and mask augmentations are applied, before feeding to the model. Similar to our previous work \cite{DBLP:conf/bildmed/SternSRKKSWE21}, the dataset was augmented with vertical and horizontal flip, rotation of $\pm 60^{\circ}$, and affine translations with $\pm 10\%$. Additionally, image augmentations comprising of pixel shifting in range $\pm 1\%$, shearing in range $\pm0.1$, brightness in range $\pm 0.2$, contrast in range from $0.3$ to $0.5$, random saturation in range from $0.5$ to $2.0$, and hue in range of $\pm0.1$ are applied. All augmentations are applied with a probability of $50\%$.

\subsection{Experiments}
\label{experiments}
Firstly, in the intra-operative domain, a $4$-fold cross-validation is performed on dataset $A.1$. Additionally, the models are evaluated on an independent test set $A.2$. In the simulator domain, a $5$-fold cross-validation is performed on the dataset $B.1$. Our baseline results were presented in \cite{DBLP:conf/bildmed/SternSRKKSWE21} for a \textit{Gaussian} heatmap with a $\sigma_1 = 1$. Here, we recompute the baseline for $\sigma_1\in{1, 2, 3}$, with the refined labels and the data split as described in Section \ref{datasets}. We present a comparison of the model variants described in Section \ref{architecture} with the baseline. Furthermore, we perform a sensitivity analysis with different parameters of the heatmap distribution, namely \textit{Gaussian} with $\sigma_1\in{1, 2, 3, 4}$, $\sigma_2\in{1, 2, 3}$ and \textit{Tanh} with $\alpha\in{7, 10.5}$. Here, the effect on the model performance in relation to varying $\sigma_1$ and $\sigma_2$ are presented. Moreover, we present an evaluation with a localisation metric, as described in Section \ref{evaluation}. Finally, we present a comparison of the evaluation with $3$ different radii around the ground-truth point, namely $6, 8$, and $10$ pixels, for the best-performing model in each domain. The network is trained with a learning rate of $0.001$, with an \textit{Adam} optimizer. A learning rate decay scheme is used to reduce the rate by a factor of $0.1$ upon a plateau for $10$ epochs. The models were trained on one of \textit{NVIDIA Quattro P6000}, \textit{NVIDIA TITAN RTX}, or \textit{NVIDIA TITAN V}. The \textit{PyTorch} library was used to implement the model pipeline.

\section{Results and Discussion}

\subsection{Results}
\label{results}
\begin{figure*}
  \includegraphics[width=1\textwidth]{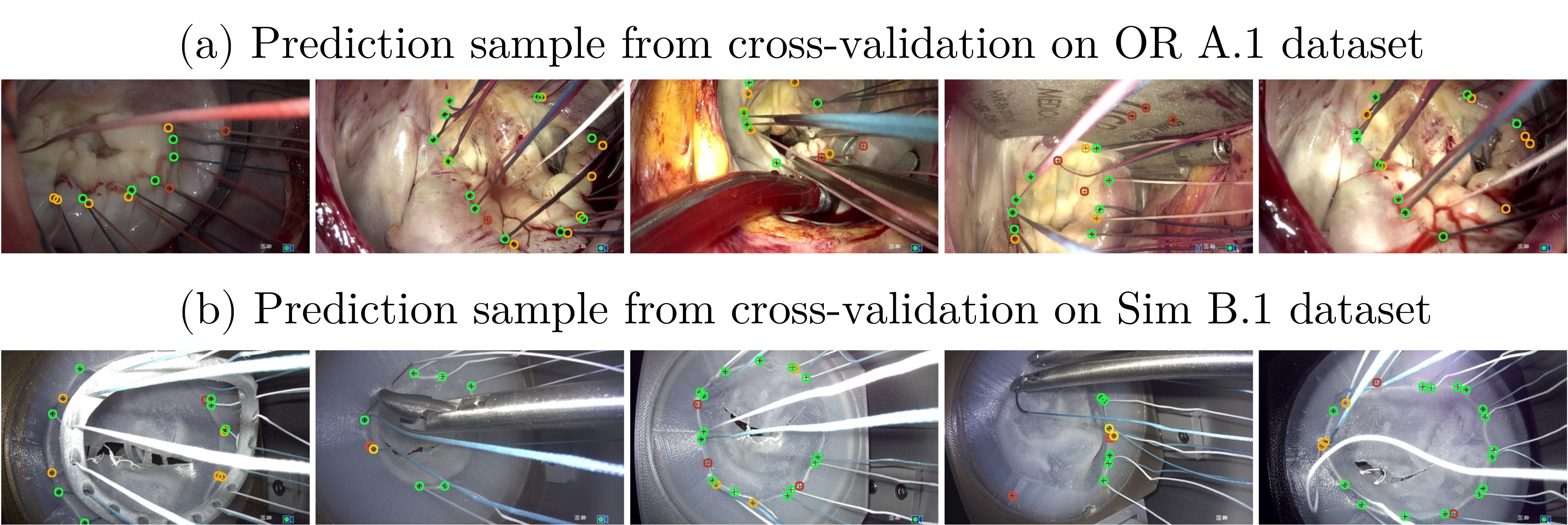}
\caption{Samples of prediction from (a) Cross-validation on the OR dataset $A.1$ (b) Cross-validation on the Sim dataset $B.1$. Legend: Green - True Positive, Orange - False Negative, Red - False Positive}
\label{fig:predictions}       
\end{figure*}

The results of the $4$-fold cross-validation in the intra-operative domain on dataset $A.1$ are presented in Table \ref{tab:results_overview} (a). Additionally, an evaluation of two variants of the proposed model in comparison with the baseline from our previous work \cite{DBLP:conf/bildmed/SternSRKKSWE21}, on the intra-operative test set $A.2$ is shown in Table \ref{tab:results_overview} (b). In the simulator domain, results of the $5$-fold cross-validation are presented in Table \ref{tab:results_overview} (c). Samples from prediction from cross-validation in the intra-operative ($A.1$) and the simulator domain ($B.1$) are shown in Fig. \ref{fig:predictions}.

Firstly, from the cross-validation in the intra-operative dataset ($A.1$), it can be seen that in comparison with the our previous baseline with the value of $\sigma_1=1$ \cite{DBLP:conf/bildmed/SternSRKKSWE21}, the performance of the model increases with $\sigma_1=2$. For both the values of $\sigma_1=2, \sigma_1=3$, the model performs better than while using a \textit{Tanh} distribution, with respective values of $\alpha=7, \alpha=10.5$ (mean $F_1$ $+0.0082$ for $\sigma_1=3, \alpha=10.5$ OR $A.1$ c.f. Table \ref{tab:results_overview} (a)). To recall, $\sigma_1$ here denotes the spread of the \textit{Gaussian} distribution used to create the masks. $\sigma_2$ refers to the parameters of the local differentiable \textit{Gaussian} layer used in the proposed model variants.

In the intra-operative dataset, the Variant $1$ of the proposed model with $\sigma_1=3$ outperforms the baseline from our previous work \cite{DBLP:conf/bildmed/SternSRKKSWE21} with $\sigma_1=1$ (mean $F_1$ $+0.0082$ for $\sigma_1=3$ OR $A.1$ c.f. Table \ref{tab:results_overview} (a)). Variant $1$ also outperforms the baseline model with the same $\sigma_1$ value of $3$ (mean $F_1$ $+0.0080$ for $\sigma_1=3$ OR $A.1$ c.f. Table \ref{tab:results_overview} (a)). In this case, the difference is the differentiable local \textit{Gaussian} and \textit{SoftArgMax} layers in the model architecture. A larger spread of the \textit{Gaussian} distribution provides more likelihood values around every landmark and additionally reduces the imbalance of the pixels in the dataset, thereby helping the model learn better. However, a larger spread around the suture point also means that the model is prone to confounding from nearby points due to overlapping distributions. Similarly, in the Simulator domain, the proposed model Variant $1$ with $\sigma_1=2$ outperforms the baseline from our previous work \cite{DBLP:conf/bildmed/SternSRKKSWE21} (mean $F_1$ $+0.0865$ Sim $B.1$ c.f. Table \ref{tab:results_overview} (c)), with $\sigma=1$, and the baseline model with the same value of $\sigma_1=2$ (mean $F_1$ $+0.0354$ Sim $B.1$ c.f. Table \ref{tab:results_overview} (c)).

\begin{table}[htbp]
\centering
\resizebox{\linewidth}{!}{
\begin{tabular}{lllll} \hline
\multicolumn{5}{l}{(a) Cross-validation on $OR$ data ($A.1$)} (Higher is better)\\ \hline
  Experiment & Mask distribution              & $PPV$                 &  $TPR$                & $F_1$               \\ \hline
  Baseline \cite{DBLP:conf/bildmed/SternSRKKSWE21}  & Gauss, $\sigma =1$   & $62.2700 \pm 9.54$  &  $33.1350 \pm 6.68$  & $0.4376 \pm 0.06$ \\ 
  Baseline \cite{DBLP:conf/bildmed/SternSRKKSWE21} & Gauss, $\sigma =2$   & $64.4450 \pm 11.01$ &  $37.6525 \pm 8.46$  & $0.4720 \pm 0.05$ \\ 
  Baseline \cite{DBLP:conf/bildmed/SternSRKKSWE21}  & Gauss, $\sigma =3$   & $65.8775 \pm 8.95$  &  $37.8200 \pm 10.58$  & $0.4718 \pm 0.08$ \\
  Baseline \cite{DBLP:conf/bildmed/SternSRKKSWE21}  & Tanh,  $\alpha =7$   & $69.2850 \pm 6.42$  &  $34.6900 \pm 5.74$  & $0.4494 \pm 0.05$ \\
  Baseline \cite{DBLP:conf/bildmed/SternSRKKSWE21}   & Tanh,  $\alpha =10.5$ & $68.3550 \pm 7.90$  &  $35.2900 \pm 5.55$  & $0.4636 \pm 0.06$ \\
  Variant 1  & Gauss, $\sigma_1 =2$ & $68.8400 \pm 7.84$  &  $38.0650 \pm 9.30$  & $0.4789 \pm 0.06$ \\ 
  Variant 1  & Gauss, $\sigma_1 =3$ & $67.2100 \pm 11.08$ &  $39.5225 \pm 10.23$ & $\mathbf{0.4798 \pm 0.04}$ \\ 
  Variant 1  & Gauss, $\sigma_1 =4$ & $67.3275 \pm 16.92$ &  $29.3600 \pm 7.77$  & $0.3711 \pm 0.04$ \\
  Variant 2  & Gauss, $\sigma_1 =2$ & $74.7625 \pm 8.30$  &  $33.9950 \pm 9.99$  & $0.4576 \pm 0.09$ \\ \hline
  
  \multicolumn{5}{l}{(b) Results on additional $OR$ data test set ($A.2$)} \\ \hline
  Experiment & Mask distribution          & $PPV$                 &  $TPR$                & $F_1$               \\ \hline
  Baseline \cite{DBLP:conf/bildmed/SternSRKKSWE21}  & Gauss, $\sigma =2$   & $76.0150 \pm 7.36$  &  $28.5025 \pm 2.53$ & $0.4126 \pm 0.02$ \\ 
  Baseline \cite{DBLP:conf/bildmed/SternSRKKSWE21}  & Gauss, $\sigma =3$   & $72.6550 \pm 1.43$  &  $28.9350 \pm 1.14$  & $0.4138 \pm 0.01$ \\
  Baseline \cite{DBLP:conf/bildmed/SternSRKKSWE21}  & Tanh,  $\alpha =7$   & $76.2525 \pm 3.52$  &  $26.3000 \pm 2.20$ & $0.3902 \pm 0.02$ \\ 
  Baseline \cite{DBLP:conf/bildmed/SternSRKKSWE21}  & Tanh,  $\alpha =10.5$ & $74.1375 \pm 5.83$  &  $28.2450 \pm 2.18$  & $0.4084 \pm 0.03$ \\
  Variant 1  & Gauss, $\sigma_1 =2$ & $76.6475 \pm 5.31$  &  $28.0750 \pm 2.92$ & $0.4086 \pm 0.02$ \\ 
  Variant 1  & Gauss, $\sigma_1 =3$ & $73.3200 \pm 2.94$  &  $30.4550 \pm 3.76$ & $\mathbf{0.4290 \pm 0.04}$ \\ 
  Variant 1  & Gauss, $\sigma_1 =4$ & $67.5825 \pm 8.42$  &  $22.3425 \pm 3.05$ & $0.3317 \pm 0.03$ \\ 
  Variant 2  & Gauss, $\sigma_1 =2$ & $78.3350 \pm 2.88$  &  $26.3875 \pm 1.79$ & $0.3945 \pm 0.02$ \\ \hline
  
  \multicolumn{5}{l}{(c) Cross-validation on Simulator data ($B.1$)} \\ \hline
  Experiment & Mask distribution             & $PPV$                 &  $TPR$                & $F_1$               \\ \hline
  Baseline \cite{DBLP:conf/bildmed/SternSRKKSWE21}  & Gauss, $\sigma =1$   & $78.3260 \pm 4.44$  &  $61.6360 \pm 7.96$ & $0.6869 \pm 0.06$ \\ 
  Baseline \cite{DBLP:conf/bildmed/SternSRKKSWE21}  & Gauss, $\sigma =2$   & $81.3900 \pm 5.18$  &  $67.9540 \pm 9.82$ & $0.7380 \pm 0.08$ \\ 
  Variant 1  & Gauss, $\sigma_1 =2$ & $83.2840 \pm 3.76$  &  $72.4860 \pm 8.56$ & $\mathbf{0.7734 \pm 0.06}$ \\ 
  Variant 1  & Gauss, $\sigma_1 =3$ & $78.4560 \pm 6.38$  &  $64.6160 \pm 9.72$ & $0.7057 \pm 0.07$ \\ 
  Variant 2  & Gauss, $\sigma_1 =2$ & $81.0100 \pm 5.92$  &  $73.7160 \pm 7.34$ & $0.7704 \pm 0.06$ \\ \hline
  \end{tabular}%
}
\caption{Results of baselines and model variants on (a) OR Cross-validation dataset $A.1$, (b) OR Test dataset $A.2$, (c) Sim Cross-validation dataset $B.1$}
\label{tab:results_overview}
\end{table}

In the intra-operative domain, Variant $2$ of the proposed model does not outperform the baseline with the corresponding $\sigma_1$ value (mean $F_1$ $-0.0144$ for $\sigma_1=3$ OR $A.1$ c.f. Table \ref{tab:results_overview} (a)). In the simulator domain however, the Variant $2$ outperforms the corresponding baseline (mean $F_1$ $+0.0324$ Sim $B.1$ c.f. Table \ref{tab:results_overview} (c)). Binary masks in this case, without a likelihood distribution constitute a highly imbalanced dataset, which hampers the learning process and affects performance. In both domains, the model Variant $1$ yields the best performing model.

\begin{figure*}
  \includegraphics[width=1\textwidth]{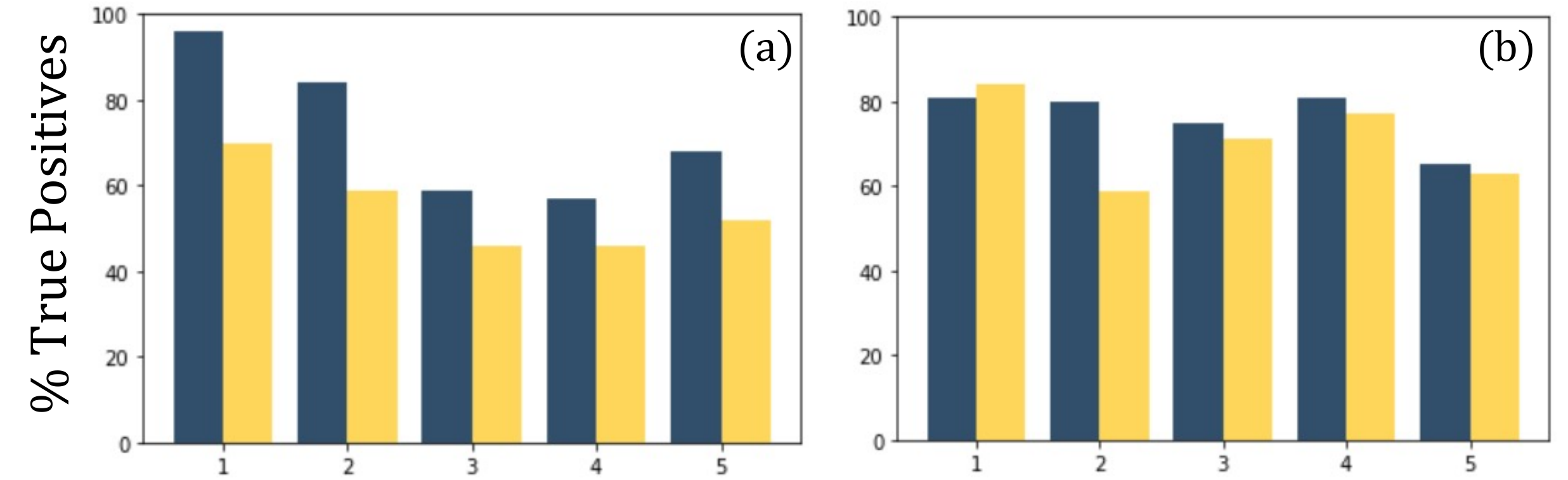}
\caption{A comparison of the percentage of True Positives detected in each fold in the simulator domain cross-validation dataset $B.1$. Light blue bars denote the model with $\sigma_1=2$, $\sigma_2=1$; Orange bars denote the model with with $\sigma_1=3$, $\sigma_2=1$; (a) provides a comparison of the subset containing points close to each other. (b) subset not close to each other.}
\label{fig:tp_comparison}       
\end{figure*}

Furthermore, for values $\sigma_1 = 2$, $\sigma_1 = 3$, the values of $\sigma_2$ are varied between $1, 2$, and $3$ and the results are presented in Table \ref{tab:gauss_comparison}. In each domain, the best performing model is with the value $\sigma_2 = 1$. In both the cases of intra-operative and the simulator domains, there is a best-performing value of ($\sigma_1$, $\sigma_2$) after which the performance of the model drops. In the case of the intra-operative domain, this performance occurs at ($\sigma_1=3$, $\sigma_2=1$) and in the case of the simulator domain, at ($\sigma_1=2$, $\sigma_2=1$). This is due to the trade-off that occurs while increasing the spread of the distribution around the suture points. In order to understand this trade-off, the model performance is analysed at the level of two different subsets. Firstly, a subset of close-points are defined as the points that are within a distance of $15$ pixels within each other. The rest of the points are categorised as non-close points. Then, the change in the \textit{True Positive} points, as we go from $\sigma_1=2$ to $\sigma_1=3$ is analysed. An example illustration in the simulator domain is shown in Figure \ref{fig:tp_comparison}. It can be seen that the drop in the percentage of \textit{True Positives} is higher in the case of the close subset in comparison to the points that are not located close to each other.

\begin{table}[htbp]
\centering
\resizebox{\linewidth}{!}{
\begin{tabular}{llllll} \hline 
\multicolumn{6}{l}{(a) Cross-validation on OR dataset ($A.1$)} (Higher is better) \\ \hline
  Metric & Experiment & Mask distribution                & $\sigma_2=1$        & $\sigma_2=2$        & $\sigma_2=3$     \\ \hline
    PPV  & Variant 1  & Gauss, $\sigma_1 =2$ & $68.8400 \pm 7.84$  & $\mathbf{75.1825 \pm 6.42}$  & $68.3875 \pm 6.90$   \\  
         & Variant 1  & Gauss, $\sigma_1 =3$ & $67.2100 \pm 11.08$ & $70.6550 \pm 13.90$ & $63.4375 \pm 13.53$  \\ \hline
         
    TPR  & Variant 1  & Gauss, $\sigma_1 =2$ & $38.0650 \pm 9.306$ & $35.4575 \pm 6.26$  & $35.5925 \pm 9.03$        \\ 
         & Variant 1  & Gauss, $\sigma_1 =3$ & $\mathbf{39.5225 \pm 10.23}$ & $33.9025 \pm 3.74$  & $35.8525 \pm 8.46$ \\ \hline
         
  $F_1$  & Variant 1  & Gauss, $\sigma_1 =2$ & $0.4789 \pm 0.06$ & $0.4781 \pm 0.06$ & $0.4582 \pm 0.07$ \\  
         & Variant 1  & Gauss, $\sigma_1 =3$ & $\mathbf{0.4798 \pm 0.04}$ & $0.4503 \pm 0.03$ & $0.4485 \pm 0.07$ \\  \hline
         
\multicolumn{6}{l}{(b) Cross-validation on Simulator dataset ($B.1$)} \\ \hline
  Metric & Experiment & Mask distribution                & $\sigma_2=1$      & $\sigma_2=2$      & $\sigma_2=3$     \\ \hline
    PPV  & Variant 1  & Gauss, $\sigma_1 =2$ & $\mathbf{83.2840 \pm 3.76}$  & $80.8760 \pm 7.53$  & $82.0220 \pm 5.14$   \\  
         & Variant 1  & Gauss, $\sigma_1 =3$ & $78.4560 \pm 6.38$  & $77.8520 \pm 9.46$  & $79.6720 \pm 4.77$  \\ \hline
         
    TPR  & Variant 1  & Gauss, $\sigma_1 =2$ & $\mathbf{72.4860 \pm 8.56}$  & $68.0740 \pm 5.00$  & $64.9800 \pm 9.81$        \\ 
         & Variant 1  & Gauss, $\sigma_1 =3$ & $64.6160 \pm 9.72$  & $66.1020 \pm 11.25$ & $67.2180 \pm 10.76$ \\ \hline
         
  $F_1$  & Variant 1  & Gauss, $\sigma_1 =2$ & $\mathbf{0.7734 \pm 0.06}$   & $0.7368 \pm 0.05$ & $0.7188 \pm 0.06$ \\  
         & Variant 1  & Gauss, $\sigma_1 =3$ & $0.7057 \pm 0.07$   & $0.7116 \pm 0.09$ & $0.7254 \pm 0.07$ \\  \hline
  \end{tabular}%
}
\caption{Comparison of different \textit{Gaussian} values used for creating the suture masks ($\sigma_1$) vs. the \textit{Gaussian} values used in the local differentiable \textit{Gaussian} layer ($\sigma_2$); on the OR ($A.1$) and Simulator dataset ($B.1$).}
\label{tab:gauss_comparison}
\end{table}

Moreover, we compute the root mean square error of the Euclidean distance as explained in Section \ref{evaluation}, the results of which are presented in Table \ref{tab:metric_distance}. As can be seen from Table \ref{tab:metric_distance}, the results are different as compared to the $F_1$ score metric presented in Table \ref{tab:gauss_comparison}. Although the $RMSE$ distance provides an indication of the closeness of the points to the ground truth labels, it is difficult to analyse a case where the $RMSE$ of two models are the same despite one of the models predicting more False Positives, since the metric is averaged over each predicted point. An example of this is shown in Figure \ref{fig:evaluation} (b). Finally, we present an evaluation with $3$ different radii around the ground-truth point for which a match is allocated, namely $6$ pixels, $8$ pixels, and $10$ pixels, for the best-performing model in each domain, as can be seen in Table \ref{tab:threshold_comparison}.

\begin{table}[htbp]
\centering
\resizebox{\linewidth}{!}{
\begin{tabular}{lllll} \hline 
  \multicolumn{5}{l}{(a) Cross-validation on OR dataset ($A.1$)} (Lower is better) \\ 
  \multicolumn{5}{l}{(Computed for $2,187$ out of $2,376$ images)}\\ \hline
    Experiment & Mask Distribution    & $\sigma_2=1$        & $\sigma_2=2$        & $\sigma_2=3$     \\ \hline
    Variant 1  & Gauss, $\sigma_1 =2$ & $22.99 \pm 6.61$  & $\mathbf{17.20 \pm 4.26}$  & $28.84 \pm 11.12$   \\  
    Variant 1  & Gauss, $\sigma_1 =3$ & $25.01 \pm 12.84$ & $24.82 \pm 17.65$ & $28.55 \pm 13.96$  \\ \hline
         
  \multicolumn{5}{l}{(b) Results on additional $OR$ data test set ($A.2$)} \\ 
  \multicolumn{5}{l}{(Computed for $391$ out of $500$ images)}\\ \hline
    Experiment & Mask distribution                & $\sigma_2=1$        & $\sigma_2=2$        & $\sigma_2=3$     \\ \hline
    Variant 1  & Gauss, $\sigma_1 =2$ & $17.2489 \pm 3.00$  & $17.3576 \pm 4.65$  & $17.4979 \pm 2.01$   \\  
    Variant 1  & Gauss, $\sigma_1 =3$ & $19.6230 \pm 4.25$  & $\mathbf{17.1994 \pm 3.67}$  & $24.5889 \pm 2.73$  \\ \hline
         
  \multicolumn{5}{l}{(c) Cross-validation on Simulator data ($B.1$)} \\ 
  \multicolumn{5}{l}{(Computed for $2,678$ out of $2,708$ images)}\\ \hline
    Experiment & Mask distribution               & $\sigma_2=1$        & $\sigma_2=2$           & $\sigma_2=3$     \\ \hline
    Variant 1  & Gauss, $\sigma_1 =2$ & $\mathbf{11.2789 \pm 6.00}$  & $13.1211 \pm 10.95$  & $11.4301 \pm 6.34$   \\  
    Variant 1  & Gauss, $\sigma_1 =3$ & $13.4286 \pm 8.49$  & $14.1760 \pm 10.45$  & $13.8515 \pm  6.70$  \\ \hline
  \end{tabular}%
}
\caption{Comparison of the $RMSE$ distance with different \textit{Gaussian} values used for creating the suture masks ($\sigma_1$) vs. the \textit{Gaussian} values used in the local differentiable \textit{Gaussian} layer ($\sigma_2$); on the (a) OR cross-validation dataset ($A.1$) (b) additional OR test dataset ($A.2$), and the simulator cross-validation dataset ($B.1$).}
\label{tab:metric_distance}
\end{table}

\begin{table}[htbp]
\centering
\resizebox{\linewidth}{!}{
\begin{tabular}{llllll} \hline 
\multicolumn{6}{l}{(a) Cross-validation on $OR$ data ($A.1$)} (Higher is better)\\ \hline
  Experiment & Mask distribution    & Radius    & $PPV$               &  $TPR$               & $F_1$               \\ \hline
  Variant 1  & Gauss, $\sigma_1 =3$ & $6$px     & $67.2100 \pm 11.0821$ &  $39.5225 \pm 10.2259$  & $0.4798 \pm 0.0427$ \\ 
  Variant 1  & Gauss, $\sigma_1 =3$ & $8$px     & $69.1575 \pm 10.80$ &  $39.5225 \pm 10.23$ & $0.4946 \pm 0.05$ \\ 
  Variant 1  & Gauss, $\sigma_1 =3$ & $10$px    & $\mathbf{70.0500 \pm 10.59}$ &  $\mathbf{41.3800 \pm 11.14}$  & $\mathbf{0.5014 \pm 0.05}$ \\ \hline
  
  \multicolumn{6}{l}{(b) Results on additional $OR$ data test set ($A.2$)} \\ \hline
  Experiment & Mask distribution    & Radius    & $PPV$               &  $TPR$               & $F_1$               \\ \hline
  Variant 1  & Gauss, $\sigma_1 =3$ & $6$px     & $73.3200 \pm 2.94$ &  $30.4550 \pm 3.76$  & $0.4290 \pm 0.04$ \\ 
  Variant 1  & Gauss, $\sigma_1 =3$ & $8$px     & $75.7525 \pm 2.68$ &  $30.4550 \pm 3.76$ & $0.4433 \pm 0.04$ \\ 
  Variant 1  & Gauss, $\sigma_1 =3$ & $10$px    & $\mathbf{76.4100 \pm 2.86}$ &  $\mathbf{31.74 \pm 3.91}$  & $\mathbf{0.4471 \pm 0.04}$ \\ \hline 
  
  \multicolumn{6}{l}{(c) Cross-validation on Simulator data ($B.1$)} \\ \hline
  Experiment & Mask distribution    & Radius.   & $PPV$              &  $TPR$              & $F_1$               \\ \hline
  Variant 1  & Gauss, $\sigma_1 =2$ & $6$px     & $83.2840 \pm 3.76$ &  $72.4860 \pm 8.56$  & $0.7734 \pm 0.06$ \\ 
  Variant 1  & Gauss, $\sigma_1 =2$ & $8$px     & $84.0475 \pm 3.95$ &  $69.9425 \pm 7.70$  & $0.7689 \pm 0.06$ \\ 
  Variant 1  & Gauss, $\sigma_1 =2$ & $10$px    & $\mathbf{85.1800 \pm 4.06}$ &  $\mathbf{72.0200 \pm 7.46}$  & $\mathbf{0.7790 \pm 0.05}$ \\ \hline
  \end{tabular}%
}
\caption{Comparison of evaluation with three different radii around the ground-truth point, for the best performing models on (a) OR cross-validation dataset $A.1$, (b) additional OR Test dataset $A.2$, (c) Simulator cross-validation dataset $B.1$}
\label{tab:threshold_comparison}
\end{table}


\subsection{Discussion}
\label{discussion}

In this paper, as an extension to our previous work \cite{DBLP:conf/bildmed/SternSRKKSWE21}, we tackle the suture detection task by introducing a differentiable $2D$ \textit{Gaussian} filter layer, and an additional differentiable convolutional $2D$ spatial convolutional \textit{Soft-Argmax} layer. Unlike other works \cite{iqbal_hand_2018,Chandran2020AttentionDrivenCF}, that use a \textit{Soft-Argmax} layer to directly extract the landmarks from the heatmap from a single channel, we present its use as a form of local non-maximum suppression to filter out points with low likelihood of being a suture. Firstly, we perform experiments comparing the baseline from our previous work \cite{DBLP:conf/bildmed/SternSRKKSWE21}, with different values of $\sigma_1$. Here, we also present comparison of the \textit{Gaussian} distribution with a \textit{Tanh} distribution with a similar spread. Then we present two variants of our proposed model in comparison with the baseline (c.f. Table \ref{tab:results_overview}). Further, we present experiments by varying values of $\sigma_1\in1, 2, 3, 4$ and $\sigma_2\in1, 2, 3$ (c.f. Table \ref{tab:gauss_comparison}). In addition to the evaluation with the $F_1$ score, we compute an $RMSE$ metric (c.f. Table \ref{tab:metric_distance}). The $RMSE$ metric has a limitation in comparing the models while taking into account the False Positives, as explained in Section \ref{evaluation}. In the intra-operative domain, the Variant $1$ with values ($\sigma_1=3, \sigma_2=1$) is the best performing model with an $F_1$ score of $0.4798\pm0.04$ OR $A.1$, $0.4290\pm0.04$ OR $A.2$ c.f. Table \ref{tab:results_overview} (a) and (b), and Variant $1$ with values ($\sigma_1=2, \sigma_2=1$) is the best performing model with an $F_1$ score of $0.7734\pm0.06$ Simulator $B.1$, c.f. Table \ref{tab:results_overview} (c). The intra-operative dataset is a highly heterogeneous dataset comprising of images from different viewing angles, scale, light sources, and white balance. Furthermore, the intra-operative datasets contain endoscopic artefacts caused due to specularities, and occlusions from tissue or surgical instruments in the scene which make it a challenging dataset to learn from. Finally, it is often the case that two sutures are stitched close to each other. This makes it further difficult for the model, and a human reader, to distinguish nearby sutures. In particular, the final $2D$ \textit{Gaussian} filter layer and the convolutional $2D$ spatial \textit{Soft-Argmax} layer operate locally with a window and are prone to be confounded by closely occurring suture points. This is especially true in the case of higher \textit{Gaussian} $\sigma_1$ values, as can be seen in Table \ref{tab:gauss_comparison}. Varying the values $\sigma_1$ and $\sigma_2$ each have an effect on model performance in relation to the number of points in the dataset that are nearby or farther away from each other. In this regard, an adaptive variation in the \textit{Gaussian} distribution is a potential future work, to handle these variations.

Besides providing quantitative information for analysis of endoscopic data, the learned representations from the suture detection task can also be used to support other learning objectives. In particular, this task is relevant in the context of generative models to transform data from the simulator to the intra-operative domain \cite{Engelhardt_Miccai18,engelhardt_cross-domain_2019}. In our recent work \cite{9496194}, we show that suture detection models can be used to mutually improve generative domain transformation in endoscopy. 

\section{Conclusion}
\label{conclusion}

In this work, we tackle the task of suture detection for endoscopic images by formulating it as a multi-instance sparse heatmap-regression problem. We extend our previous work \cite{DBLP:conf/bildmed/SternSRKKSWE21} and improve upon the previously reported baselines. We introduce a novel \textit{Gaussian} filter layer and a differentiable convolutional \textit{Soft-Argmax} layers. We compare multiple distribution functions and present two variants of the model that outperform the baselines. Suture detection is an important task that can further be used towards supporting the learning of related objectives for endoscopic image analysis.

\textbf{Conflict of interest:} All authors declare that they have no conflict of interest.
\textbf{Funding:} The research was supported by the German Research Foundation DFG Project 398787259, DE 2131/2-1 and EN 1197/2-1 and by Informatics for Life funded by the Klaus Tschira Foundation
\textbf{Availability of data and material:} The data is released within the scope of the AdaptOR2021 MICCAI challenge at \url{https://adaptor2021.github.io/}.
\textbf{Ethics Approval:} The  study  was  approved  by  the  Local  Ethics  Commitee from  University  Hospital  Heidelberg.  Registration  numbers are  S-658/2016  (12.09.2017)  and  S-777/2019  (20.11.2019). The study was performed in accordance with the ethical standards as laid down in the 1964 Declaration of Helsinki and its later amendments or comparable ethical standards. \textbf{Consent to participate} Informed consent was obtained from all individual participants included in the study. \textbf{Consent for publication} Not applicable
\textbf{Code availability} The authors plan to make the code publicly available in the subsequent months. 

\bibliographystyle{spmpsci}      
\bibliography{references.bib} 

\begin{thebibliography}{10}
\providecommand{\url}[1]{{#1}}
\providecommand{\urlprefix}{URL }
\expandafter\ifx\csname urlstyle\endcsname\relax
  \providecommand{\doi}[1]{DOI~\discretionary{}{}{}#1}\else
  \providecommand{\doi}{DOI~\discretionary{}{}{}\begingroup
  \urlstyle{rm}\Url}\fi

\bibitem{brosch_deep_2015}
Brosch, T., Yoo, Y., Tang, L.Y.W., Li, D.K.B., Traboulsee, A., Tam, R.: Deep
  {Convolutional} {Encoder} {Networks} for {Multiple} {Sclerosis} {Lesion}
  {Segmentation}.
\newblock In: N.~Navab, J.~Hornegger, W.M. Wells, A.F. Frangi (eds.) Medical
  {Image} {Computing} and {Computer}-{Assisted} {Intervention} – {MICCAI}
  2015, Lecture {Notes} in {Computer} {Science}, pp. 3--11. Springer
  International Publishing, Cham (2015).
\newblock \doi{https://doi.org/10.1007/978-3-319-24574-4_1}

\bibitem{bulat_human_2016}
Bulat, A., Tzimiropoulos, G.: Human {Pose} {Estimation} via {Convolutional}
  {Part} {Heatmap} {Regression}.
\newblock In: B.~Leibe, J.~Matas, N.~Sebe, M.~Welling (eds.) Computer {Vision}
  – {ECCV} 2016, Lecture {Notes} in {Computer} {Science}, pp. 717--732.
  Springer International Publishing, Cham (2016).
\newblock \doi{https://doi.org/10.1007/978-3-319-46478-7_44}

\bibitem{carpentier_new_1971}
Carpentier, A., Deloche, A., Dauptain, J., Soyer, R., Blondeau, P., Piwnica,
  A., Dubost, C., McGoon, D.C.: A new reconstructive operation for correction
  of mitral and tricuspid insufficiency.
\newblock The Journal of Thoracic and Cardiovascular Surgery \textbf{61}(1),
  1--13 (1971)

\bibitem{casselman_filip_p._mitral_2003}
{Casselman Filip P.}, {Van Slycke Sam}, {Wellens Francis}, {De Geest Raphael},
  {Degrieck Ivan}, {Van Praet Frank}, {Vermeulen Yvette}, {Vanermen Hugo}:
  Mitral {Valve} {Surgery} {Can} {Now} {Routinely} {Be} {Performed}
  {Endoscopically}.
\newblock Circulation \textbf{108}(10\_suppl\_1), II--48 (2003).
\newblock \doi{https://doi.org/10.1161/01.cir.0000087391.49121.ce}

\bibitem{Chandran2020AttentionDrivenCF}
Chandran, P., Bradley, D., Gross, M., Beeler, T.: Attention-driven cropping for
  very high resolution facial landmark detection.
\newblock 2020 IEEE/CVF Conference on Computer Vision and Pattern Recognition
  (CVPR) pp. 5860--5869 (2020).
\newblock \doi{https://doi.org/10.1109/CVPR42600.2020.00590}

\bibitem{duffner_connexionist_2005}
Duffner, S., Garcia, C.: A connexionist approach for robust and precise facial
  feature detection in complex scenes.
\newblock In: {ISPA} 2005. {Proceedings} of the 4th {International} {Symposium}
  on {Image} and {Signal} {Processing} and {Analysis}, 2005., pp. 316--321
  (2005).
\newblock \doi{https://doi.org/10.1109/ISPA.2005.195430}.
\newblock ISSN: 1845-5921

\bibitem{Engelhardt_Miccai18}
Engelhardt, S., De~Simone, R., Full, P.M., Karck, M., Wolf, I.: Improving
  surgical training phantoms by hyperrealism: deep unpaired image-to-image
  translation from real surgeries.
\newblock In: MICCAI, pp. 747--755. Springer International Publishing (2018)

\bibitem{EngelhardtMiccai14}
Engelhardt, S., De~Simone, R., Zimmermann, N., Al-Maisary, S., Nabers, D.,
  Karck, M., Meinzer, H.P., Wolf, I.: Augmented reality-enhanced endoscopic
  images for annuloplasty ring sizing.
\newblock In: Augmented Environments for Computer-Assisted Interventions, pp.
  128--137. Springer International Publishing (2014)

\bibitem{3093-04}
Engelhardt, S., Kolb, S., De~Simone, R., Karck, M., Meinzer, H.P., Wolf, I.:
  Endoscopic feature tracking for augmented-reality assisted prosthesis
  selection in mitral valve repair.
\newblock In: Proc {SPIE}, Medical Imaging: Image-Guided Procedures, Robotic
  Interventions, and Modeling, vol. 9786, pp. 402--408 (2016)

\bibitem{EngelhardtICVTS}
Engelhardt, S., Sauerzapf, S., Brčić, A., Karck, M., Wolf, I., De~Simone, R.:
  {Replicated mitral valve models from real patients offer training
  opportunities for minimally invasive mitral valve repair}.
\newblock Interactive CardioVascular and Thoracic Surgery \textbf{29}(1),
  43--50 (2019).
\newblock \doi{https://doi.org/10.1093/icvts/ivz008}

\bibitem{EngelhardtIJCARS2019}
Engelhardt, S., Sauerzapf, S., Preim, B., Karck, M., Wolf, I., De~Simone, R.:
  Flexible and comprehensive patient-specific mitral valve silicone models with
  chordae tendinae made from {3D}-printable molds.
\newblock Int J Comput Assist Radiol Surg \textbf{14}(7), 1177--1186 (2019)

\bibitem{engelhardt_cross-domain_2019}
Engelhardt, S., Sharan, L., Karck, M., De~Simone, R., Wolf, I.: Cross-{Domain}
  {Conditional} {Generative} {Adversarial} {Networks} for {Stereoscopic}
  {Hyperrealism} in {Surgical} {Training}.
\newblock In: {MICCAI} (2019).
\newblock \doi{https://doi.org/10.1007/978-3-030-32254-0_18}

\bibitem{fan_approaching_2016}
Fan, H., Zhou, E.: Approaching human level facial landmark localization by deep
  learning.
\newblock Image and Vision Computing \textbf{47}, 27--35 (2016).
\newblock \doi{https://doi.org/10.1016/j.imavis.2015.11.004}

\bibitem{hashemi_asymmetric_2019}
Hashemi, S.R., Salehi, S.S.M., Erdogmus, D., Prabhu, S.P., Warfield, S.K.,
  Gholipour, A.: Asymmetric {Loss} {Functions} and {Deep} {Densely} {Connected}
  {Networks} for {Highly} {Imbalanced} {Medical} {Image} {Segmentation}:
  {Application} to {Multiple} {Sclerosis} {Lesion} {Detection}.
\newblock IEEE access : practical innovations, open solutions \textbf{7},
  721--1735 (2019).
\newblock \doi{https://doi.org/10.1109/ACCESS.2018.2886371}

\bibitem{DBLP:journals/cmpb/HervellaRNPO20}
Hervella, {\'{A}}.S., Rouco, J., Novo, J., Penedo, M.G., Ortega, M.: Deep
  multi-instance heatmap regression for the detection of retinal vessel
  crossings and bifurcations in eye fundus images.
\newblock Comput. Methods Programs Biomed. \textbf{186}, 105201 (2020).
\newblock \doi{https://doi.org/10.1016/j.cmpb.2019.105201}

\bibitem{iqbal_hand_2018}
Iqbal, U., Molchanov, P., Breuel, T., Gall, J., Kautz, J.: Hand {Pose}
  {Estimation} via {Latent} 2.{5D} {Heatmap} {Regression}.
\newblock ECCV  (2018).
\newblock \doi{https://doi.org/10.1007/978-3-030-01252-6_8}

\bibitem{kowalski_deep_2017}
Kowalski, M., Naruniec, J., Trzcinski, T.: Deep {Alignment} {Network}: {A}
  {Convolutional} {Neural} {Network} for {Robust} {Face} {Alignment}.
\newblock In: 2017 {IEEE} {Conference} on {Computer} {Vision} and {Pattern}
  {Recognition} {Workshops} ({CVPRW}), pp. 2034--2043 (2017).
\newblock \doi{https://doi.org/10.1109/CVPRW.2017.254}.
\newblock ISSN: 2160-7516

\bibitem{payer_integrating_2019}
Payer, C., Štern, D., Bischof, H., Urschler, M.: Integrating spatial
  configuration into heatmap regression based {CNNs} for landmark localization.
\newblock Medical Image Analysis \textbf{54}, 207--219 (2019).
\newblock \doi{https://doi.org/10.1016/j.media.2019.03.007}

\bibitem{eriba2019kornia}
Riba, E., Mishkin, D., Ponsa, D., Rublee, E., Bradski, G.: Kornia: an open
  source differentiable computer vision library for pytorch.
\newblock In: Winter Conference on Applications of Computer Vision (2020)

\bibitem{ronneberger_u-net_2015}
Ronneberger, O., Fischer, P., Brox, T.: U-{Net}: {Convolutional} {Networks} for
  {Biomedical} {Image} {Segmentation}.
\newblock In: N.~Navab, J.~Hornegger, W.M. Wells, A.F. Frangi (eds.) Medical
  {Image} {Computing} and {Computer}-{Assisted} {Intervention} – {MICCAI}
  2015, Lecture {Notes} in {Computer} {Science}, pp. 234--241. Springer
  International Publishing, Cham (2015).
\newblock \doi{https://doi.org/10.1007/978-3-319-24574-4_28}

\bibitem{sharan_domain_2020}
Sharan, L., Burger, L., Kostiuchik, G., Wolf, I., Karck, M., De~Simone, R.,
  Engelhardt, S.: Domain gap in adapting self-supervised depth estimation
  methods for stereo-endoscopy.
\newblock Current Directions in Biomedical Engineering \textbf{6}(1) (2020).
\newblock \doi{https://doi.org/10.1515/cdbme-2020-0004}.
\newblock Publisher: De Gruyter Section: Current Directions in Biomedical
  Engineering

\bibitem{9496194}
Sharan, L., Romano, G., Koehler, S., Kelm, H., Karck, M., De~Simone, R.,
  Engelhardt, S.: Mutually improved endoscopic image synthesis and landmark
  detection in unpaired image-to-image translation.
\newblock IEEE Journal of Biomedical and Health Informatics pp. 1--1 (2021).
\newblock \doi{10.1109/JBHI.2021.3099858}

\bibitem{sofka_fully_2017}
Sofka, M., Milletari, F., Jia, J., Rothberg, A.: Fully {Convolutional}
  {Regression} {Network} for {Accurate} {Detection} of {Measurement} {Points}.
\newblock In: M.J. Cardoso, T.~Arbel, G.~Carneiro, T.~Syeda-Mahmood, J.M.R.
  Tavares, M.~Moradi, A.~Bradley, H.~Greenspan, J.P. Papa, A.~Madabhushi, J.C.
  Nascimento, J.S. Cardoso, V.~Belagiannis, Z.~Lu (eds.) Deep {Learning} in
  {Medical} {Image} {Analysis} and {Multimodal} {Learning} for {Clinical}
  {Decision} {Support}, Lecture {Notes} in {Computer} {Science}, pp. 258--266.
  Springer International Publishing, Cham (2017).
\newblock \doi{https://doi.org/10.1007/978-3-319-67558-9_30}

\bibitem{DBLP:conf/bildmed/SternSRKKSWE21}
Stern, A., Sharan, L., Romano, G., Koehler, S., Karck, M., De~Simone, R., Wolf,
  I., Engelhardt, S.: Heatmap-based 2d landmark detection with a varying number
  of landmarks.
\newblock In: C.~Palm, T.M. Deserno, H.~Handels, A.~Maier, K.H. Maier{-}Hein,
  T.~Tolxdorff (eds.) Bildverarbeitung f{\"{u}}r die Medizin 2021 -
  Proceedings, German Workshop on Medical Image Computing, Regensburg, March
  7-9, 2021, Informatik Aktuell, pp. 22--27. Springer (2021).
\newblock \doi{https://doi.org/10.1007/978-3-658-33198-6\_7}

\bibitem{DBLP:journals/corr/SunMX15}
Sun, P., Min, J.K., Xiong, G.: Globally tuned cascade pose regression via back
  propagation with application in 2d face pose estimation and heart
  segmentation in 3d {CT} images.
\newblock CoRR \textbf{abs/1503.08843} (2015).
\newblock \urlprefix\url{http://arxiv.org/abs/1503.08843}

\bibitem{yan_survey_2018}
Yan, Y., Naturel, X., Chateau, T., Duffner, S., Garcia, C., Blanc, C.: A survey
  of deep facial landmark detection.
\newblock In: {RFIAP}. Paris, France (2018)

\bibitem{yang_stacked_2017}
Yang, J., Liu, Q., Zhang, K.: Stacked {Hourglass} {Network} for {Robust}
  {Facial} {Landmark} {Localisation}.
\newblock In: 2017 {IEEE} {Conference} on {Computer} {Vision} and {Pattern}
  {Recognition} {Workshops} ({CVPRW}), pp. 2025--2033. IEEE, Honolulu, HI, USA
  (2017).
\newblock \doi{https://doi.org/10.1109/CVPRW.2017.253}

\bibitem{zhang_detecting_2017}
Zhang, J., Liu, M., Shen, D.: Detecting {Anatomical} {Landmarks} {From}
  {Limited} {Medical} {Imaging} {Data} {Using} {Two}-{Stage} {Task}-{Oriented}
  {Deep} {Neural} {Networks}.
\newblock IEEE Transactions on Image Processing \textbf{26}(10), 4753--4764
  (2017).
\newblock \doi{https://doi.org/10.1109/TIP.2017.2721106}.
\newblock Conference Name: IEEE Transactions on Image Processing

\bibitem{DBLP:conf/miccai/ZhangLWCYLSTCXS17}
Zhang, J., Liu, M., Wang, L., Chen, S., Yuan, P., Li, J., Shen, S.G., Tang, Z.,
  Chen, K., Xia, J.J., Shen, D.: Joint craniomaxillofacial bone segmentation
  and landmark digitization by context-guided fully convolutional networks.
\newblock In: M.~Descoteaux, L.~Maier{-}Hein, A.M. Franz, P.~Jannin, D.L.
  Collins, S.~Duchesne (eds.) Medical Image Computing and Computer Assisted
  Intervention - {MICCAI} 2017 - 20th International Conference, Quebec City,
  QC, Canada, September 11-13, 2017, Proceedings, Part {II}, \emph{Lecture
  Notes in Computer Science}, vol. 10434, pp. 720--728. Springer (2017).
\newblock \doi{https://doi.org/10.1007/978-3-319-66185-8\_81}

\end{thebibliography}

%
%

\end{document}